\documentclass[manuscript]{acmart}

\AtBeginDocument{%
  \providecommand\BibTeX{{%
    \normalfont B\kern-0.5em{\scshape i\kern-0.25em b}\kern-0.8em\TeX}}}

\begin{document}

\title{Simulations in Recommender Systems: An industry perspective}

\author{Lucas Bernardi}
\email{lucas.bernardi@booking.com}
\author{Sakshi Batra}
\email{sakshi.batra@booking.com}
\author{Cintia Alicia Bruscantini}
\email{cintia.bruscantini@booking.com}
\affiliation{%
  \institution{Booking.com}
  \city{Amsterdam}
  \state{Noord-Holland}
  \country{Netherlands}
}


\begin{abstract}

The construction of effective Recommender Systems (RS) is a complex process, mainly due to the nature of RSs which involves large scale software-systems and human interactions. Iterative development processes require deep understanding of a current \textit{baseline} as well as the ability to estimate the impact of changes in multiple variables of interest. Simulations are well suited to address both challenges and potentially leading to a high velocity construction process, a fundamental requirement in commercial contexts. Recently, there has been significant interest in RS Simulation Platforms, which allow RS developers to easily craft simulated environments where their systems can be analysed. In this work we discuss how simulations help to increase velocity, we look at the literature around RS Simulation Platforms, analyse strengths and gaps and distill a set of guiding principles for the design of RS Simulation Platforms that we believe will maximize the velocity of iterative RS construction processes.
\end{abstract}


\settopmatter{printacmref=false, printfolios=false, printccs=false}
\setcopyright{none}

\maketitle

\section{Introduction}
Recommender Systems are complex systems. They are usually composed of advanced statistical algorithms relying on huge data sets and large scale distributed software systems that need to operate in a low latency regime. Furthermore, they interact with humans, giving place to hard to foresee phenomena such as paradox of choice \cite{schwartz2004paradox} and information bubbles \cite{nguyen2014exploring}. Recommender Systems are complex systems because they try to solve hard problems in complex environments. This implies that their construction process is therefore also complex. The main objective of most industrial recommender systems is to deliver value for the user as well as commercial gains for the provider which leads to iterative processes where the recommender system is continuously improved through new algorithms, user interfaces, etc. \cite{bernardi2019150} which are typically evaluated through Online Controlled Experiments (OCEs) \cite{kohavi2013online}, in order to estimate the causal impact of a new version on all the relevant metrics including user satisfaction and commercial gains among others. One of the most challenging aspects of this process is that given a current baseline system, it is not obvious which aspect (such as diversity, relevance, etc), nor which component (such as the ranker, candidate selector, serving infrastructure, etc.) should be the target of the next iteration. A common approach to address this is to develop well articulated hypotheses about \textit{issues} (e.g. current serving approach has high latency leading to higher abandonment rate) or \textit{opportunities for improvement} (e.g. point-wise ranker produces low diversity leading to very similar recommendations) in the current version, and use existing or new data to validate them. Unfortunately in many cases this is not possible, because the available data is not appropriate, or because gathering new data is too expensive.

Once a concrete improvement opportunity is detected, the current system is modified accordingly, for example by changing a prediction algorithm, the user interface or the serving infrastructure. These are usually \textit{local} interventions with measurable local effects. But they also have (and they are expected to have) global or system-wide effects, which are a lot harder to measure since they involve the system as a whole. On top of this, there is usually tension between various variables of interest such as relevance vs latency, or diversity vs relevance or adaptivity vs user satisfaction to name a few which further increases the importance of system-wide analysis. The ability to make good and principled trade-offs is key to deliver a balanced and robust recommender system.

In short, commercial recommender systems require a high velocity construction process capable of continuously increasing the delivered value. To achieve this, we identify two important challenges:
\begin{itemize}
    \item \textbf{Opportunity Identification}: in this stage we are interested in studying the current system to identify flaws and/or improvement points which might lead to a better version of the system.
    \item \textbf{System Evaluation}: Evaluate and estimate the impact of a new version of the system.
\end{itemize}

In this work we study the role of Simulations as a tool to address them. We look at existing simulation platforms, identify gaps and propose a set of principles to maximize the velocity of the recommender system construction process.

\section{Improving the RS Construction Process with Simulations}
Simulations are good fit for addressing both the \textit{Opportunity Identification} and the \textit{System Evaluation} challenges.
\subsection{Opportunity identification}
In \textit{Opportunity identification}, given a recommender system currently deployed in production, we want to find areas of improvement by understanding it's strengths and flaws. Simulations allow us to manipulate any number of variables we consider important in order to understand their impact on system behaviour. For example, we can analyse the effect of feedback delay by creating multiple simulated environments introducing perturbations in the time it takes to observe user feedback, allowing developers to extract insights about the robustness of the current baseline to different user feedback delay regimes. Simulated environments can also be designed using modular components that can be combined and reused to produce new environments which further helps with velocity (e.g. a Delay Model can be combined with a User Preference Model or an Item Availability Model).
Counterfactual Policy Evaluation (CPE) can also be used as a mechanism to manipulate variables, but their applicability is limited since for every single intervention a new counterfactual model must be constructed. Finally Online Controlled Experiments (OCE) allow to manipulate variables that are under full control such as recommendations, available items, user interface, etc. This technique is rather simple but it is costly and does not allow the manipulation of variables out of controls, such as user preferences or market conditions. In practice CPE and OCEs are mainly applied for the System Evaluation Challenge. Hence, we believe that simulations are a fundamental method to systematically identify opportunities to improve a recommender system.


\subsection{System Evaluation}
In the \textit{System Evaluation} challenge, we are interested in the \textit{causal} effect of replacing the current baseline with a new recommender system on several variables of interest. 

The gold standard approach is Online Controlled Experiments, which provide unbiased causal estimates with strong guarantees under rather weak assumptions and with maximum \textit{reality faithfulness} since actual users are exposed to the new system. At the same time have a few strong limitations affecting the velocity of the \textit{construction process} of recommender systems, concretely:
\begin{itemize}
    \item Only a few experiments can be run concurrently,
    \item each experiment has potentially high cost,
    \item they provide limited external validity, which is of particular concern in non-stationary contexts such as the ones where recommender systems operate.
\end{itemize}
Thus, OCEs are powerful but constrain the velocity of the recommender system construction process.

At the opposite end of the spectrum, Counterfactual Policy Evaluation constructs estimates relying on data collected using the baseline system (or some other data-collection policy) \cite{gilotte2018offline}. These allow users to evaluate many systems in parallel without even deploying the new system in production and therefore avoiding the associated costs. However, they also suffer from limited external validity and they provide reasonable power only if the new policy is close to the data collection policy. More importantly, since the users are never exposed to the new system, they provide weak \textit{reality faithfulness} and require very good counterfactual models which are usually very hard to build in a non-stationary environment. Evaluating a recommender system in simulated environments allows developers to run many experiments in parallel, without the costs of exposing users to the new system. Furthermore, it enables a trade-off between complexity and \textit{reality faithfulness}: developers can trade reality faithfulness with environment complexity. OCEs and CPE are located at a fixed point of the Environment-Complexity vs Reality Faithfulness trade-off plane (Figure 1), simulations are much more flexible allowing to create environments of varying complexity and reality faithfulness. 
\par

\begin{figure}
    \centering
    \includegraphics[width=80mm]{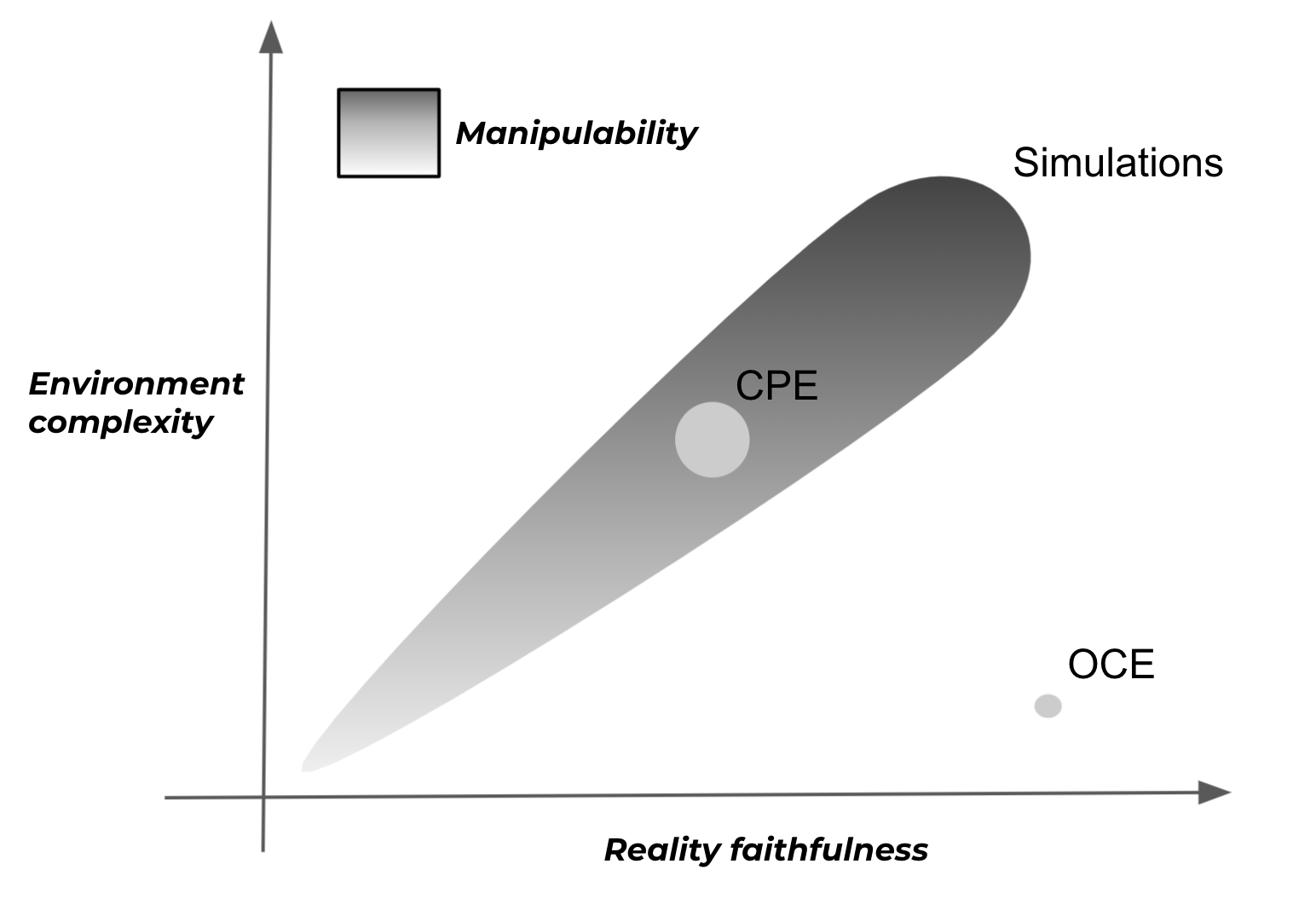}
    \caption{Comparison between Simulations, Counterfactual Policy Evaluation (CPE) and Online Controlled Experiments (OCE).}
    \label{fig:comparison_diagram}
\end{figure}

Figure \ref{fig:comparison_diagram} summarizes the trade offs between CPE, OCEs and Simulations across different dimensions. Our main observation is that unlike CPE and OCEs which are constrained to small regions of the trade-off plane, Simulations provide much larger coverage resulting in high flexibility allowing developers to strike a balance between Environment Complexity, Reality Faithfulness and Manipulability. 

\section{Simulation Platforms}

We consider the literature concerned with the development of Simulation Platforms aiming at providing a general framework for running simulations at scale. We try to understand how they help to increase the velocity of the recommender system construction process, providing our remarks for each particular case.

RecoGym \cite{rohde2018recogym} offers users with the capability to create configurable Reinforcement Learning (RL) environments for studying sequential user interaction combining \textit{organic} feedback with intermittent advertisement display, resulting in bandit feedback. While RecoGym supports sequential interaction and configuration of item/user dimensionality, it lacks flexibility in configuration of user state transitions, defining multiple reward responses such as conversions, costs, bounce off, lifetime value, etc. These characteristics can be important for real world applications. 

PyRecGym \cite{shi2019pyrecgym} extends the idea of RecoGym by allowing users to create more generic environments to accommodate a variety of recommender systems tasks instead of using synthetic data. It can support multiple input data-types and user-feedback functions. It can also ingest existing RS data-sets and use them to play out sequences of interactions between users and the RL system. PyRecGym allows flexibility in its rewards and input but its architecture makes it hard to create reusable simulated environments. Another remark is that the user feedback is simulated by replaying the data-set which might introduce bias towards the data collection policy. 

RecSim \cite{ie2019recsim} provides much more flexibility in defining the environment. It allows the creation of new environments that reflect particular aspects of user behavior (such as user preference in a document’s topic might increase/decrease over time) at different levels of abstraction. This kind of modular design encourages reusability of components across environments. This flexibility, though, also means users of the platform are required to implement layers of abstractions to emulate specific aspects of the user behaviour. Thus, making it hard to simulate a complex system and in turn potentially creating a simulation-to-reality gap.

MARS-Gym \cite{santana2020mars} is aimed more at directly using the production data, creating simulation events using it and provides evaluation metrics using CPE to overcome the bias introduced by the logging policy. It addresses the whole development pipeline: data processing directly from the production logs, model design and optimization, and multi-sided evaluation. Mars-Gym also simulates the dynamics of multi user marketplace instead of viewing each user session individually.
However, it doesn’t allow manipulating parameters in the simulation module. This lack of flexibility makes it difficult to test specific assumptions. Similar to RecoGym, MARS-Gym is limited in the kind of interactions and user feedback it supports. This also makes them unsuitable for multi-objective optimization. 

RS simulators based on logged data such as Mars-Gym, PyRecGym also suffer from bias in logged data policy and require methods to bias-correct the simulated data \cite{huang2020keeping}. Inspired by \cite{shi2019pyrecgym} Table 1 compares these platforms across several dimensions that we consider important.

\begin{table}[h]
\begin{tabular}{ |c|c|c|c|c| } 
 \hline
 & \textbf{RecoGym} & \textbf{RecSim} & \textbf{MARS-Gym} & \textbf{PyRecGym} \\
 \hline
\textit{Customized for RS} & Y & Y & Y & Y\\ 
\textit{Generalized Recommendation Task} & N & Y & N & Y\\
\textit{User Feedback Flexibility} & N & Y & N & Y\\
\textit{Marketplace Simulation} & N & N & Y & N\\
\textit{Modular Environments Support} & Y & Y & N & N\\
\textit{External real/benchmark dataset} & N & N & Y & Y\\
\hline
\end{tabular}
\caption{\label{tab:Table 1} Comparison of RS Simulation Platforms}
\end{table}
%
%

One general issue we detected is that most simulation platforms require the implementation of recommender algorithms or models for a specific run-time and complying with specific APIs which might not match the actual production run-time. This implies that the algorithms need to be re-implemented (specially the inference logic, but also the learning logic, for example in online learning such as in \cite{bernardi2020recommending}) in order to be deployed to production which clearly hurts velocity and introduces potential gaps between the \textit{simulated} version and the \textit{deployed} version.

\section{Three Principles for Building Effective Simulation Platforms}
In this section we propose a set of principles for building a industry-ready Simulation Platform:
\begin{enumerate}
    \item \textbf{Simulation code = Production code}: the implementation of a recommender system (and more specifically the algorithmic components) must be completely agnostic of the run-time environment where it runs, be it a simulated environment or a real production environment.

    \item \textbf{Minimal Input}: Simulation Platforms aid users to observe \textit{intervened} realistic environments under specific assumptions. Interventions and assumptions are specified by the user, everything else must be, as much as possible, inferred from real data.
    \item \textbf{Simulation = Reality + Assumptions + Interventions}: A Simulation Platform allows users to efficiently design simulated environments with reusable and extendable components giving developers the ability to manipulate variables and explicitly express assumptions and interventions. The Simulation Platform provides means to make good trade-offs between \textit{reality faithfulness},  \textit{environment complexity} and \textit{manipulability}. 
    
    
\end{enumerate}



\section{Future Directions}
We further think that the following directions can help to make the platform more robust and generic and ultimately help bridge the simulation-to-reality gap:
\begin{itemize}
\item \textbf{Structural Causal Models}: in light of principles 2 and 3, Structural Causal Models \cite{pearl2000causality} are an excellent framework to sit at the core of a Simulation Platform. They provide a modular language to express assumptions and interventions relying on the principle of independence of mechanisms \cite{peters2017elements}, and, to some extent they can be learned from data \cite{forre2018constraint}.

 \item \textbf{Integrated evaluation framework}: There is a need for an integrated evaluation framework which has a combination of tools which can help the developer throughout the journey, starting from doing a controlled study of specific aspects to gain understanding of algorithms, challenging assumptions by manipulating given parameter in the environment using simulator and offering capabilities to compare the desired test policy with current production policy using CPE. Once a policy has been chosen, developers should also have the capability to test the whole integrated system and deploy directly in production without much code change and eventually evaluate using OCEs. 
 
\end{itemize}

\section{Conclusions}
Simulations can be considered as a complementary tool to Online Controlled Experiments and Counterfactual Policy Evaluation. Although it can be a flexible and powerful tool for offline analysis and evaluation, a complex and realistic system might require significant development efforts. Thus, the tool needs to be generic to serve multiple use cases. Though there are some simulation platforms developed in the industry, we identified and described principles that constitute, based on our understanding, the fundamentals of an ideal simulation platform for production code development, of particular interest in the industry context.




\bibliographystyle{ACM-Reference-Format}
\bibliography{bibiliography.bib}


\end{document}